# Benchmarking Quality-Dependent and Cost-Sensitive Score-Level Multimodal Biometric Fusion Algorithms

Norman Poh, Thirimachos Bourlai, Josef Kittler, Lorene Allano, Fernando Alonso-Fernandez, Onkar Ambekar, John Baker, Bernadette Dorizzi, Omolara Fatukasi, Julian Fierrez, Harald Ganster, Javier Ortega-Garcia, Donald Maurer, Albert Ali Salah, Tobias Scheidat, and Claus Vielhauer

*Abstract*—Automatically verifying the identity of a person by means of biometrics (e.g., face and fingerprint) is an important application in our day-to-day activities such as accessing banking services and security control in airports. To increase the system reliability, several biometric devices are often used. Such a combined system is known as a multimodal biometric system. This paper reports a benchmarking study carried out within the framework of the BioSecure DS2 (Access Control) evaluation campaign organized by the University of Surrey, involving face, fingerprint, and iris biometrics for person authentication, targeting the application of physical access control in a medium-size establishment with some 500 persons. While multimodal biometrics is a well-investigated subject in the literature, there exists no benchmark for a fusion algorithm comparison. Working towards this goal, we designed two sets of experiments: quality-dependent and cost-sensitive evaluation. The quality-dependent evaluation aims at assessing how well fusion algorithms can perform under changing quality of raw biometric images principally due to change of devices. The cost-sensitive evaluation, on the other hand, investigates how well a fusion algorithm can perform given restricted computation and in the presence of software and hardware failures, resulting in errors such as failure-to-acquire and failure-to-match. Since multiple capturing devices are available, a fusion algorithm should be able to handle this nonideal but nevertheless realistic scenario. In both evaluations, each fusion algorithm is provided with scores from each biometric comparison subsystem as well as the quality measures of both the template and the query data. The response to the call of the evaluation campaign proved very encouraging, with the submission of 22 fusion systems. To the best of our knowledge, this campaign is the first attempt to benchmark quality-based multimodal fusion algorithms. In the presence of changing image quality which may be due to a change of acquisition devices and/or device capturing configurations, we observe that the top performing fusion algorithms are those that exploit automatically derived quality measurements. Our evaluation also suggests that while using all the available biometric sensors can definitely increase the fusion performance, this comes at the expense of increased cost in terms of acquisition time, computation time, the physical cost of hardware, and its maintenance cost. As demonstrated in our experiments, a promising solution which minimizes the composite cost is sequential fusion, where a fusion algorithm sequentially uses match scores until a desired confidence is reached, or until all the match scores are exhausted, before outputting the final combined score.

This work was supported by the European Union under the following projects: BioSecure (www.biosecure.info) and Mobio (www.mobioproject.org). The participating teams were supported by their respective national fund bodies: the Dutch BSIK/BRICKS project and the Spanish project TEC2006-13141-C03-03. The work of N. Poh was supported by the Swiss National Science Foundation under Advanced Researcher Fellowship PA0022_121477. The work of F. Alonso-Fernandez was supported by Consejeria de Educacion de la Comunidad de Madrid and Fondo Social Europeo. The work of J. Fierrez was supported by a Marie Curie fellowship from the European Commission.

N. Poh, J. Kittler, and O. Fatukasi are with the Centre for Vision, Speech and Signal Processing, School of Electronics and Physical Sciences, University of Surrey, Guildford, GU2 7XH, Surrey, U.K. (e-mail: normanpoh@ieee.com; j.kittler@surrey.ac.uk; o.fatukasi@surrey.ac.uk).

T. Bourlai is with the Biometrics Center, Lane Department of Computer Science and Electrical Engineering, College of Engineering and Mineral Resources, West Virginia University, Morgantown, WV 26506-6109 USA (e-mail: ThBourlai@mail.wvu.edu).

L. Allano was with Institut Telecom, Telecom and Management SudParis, 91011 Evry, France. She is currently with CEA LIST, CEA saclay-PC 72–91191 Gif-sur-Yvette Cedex, France (e-mail: lorene.allano@gmail.com).

F. Alonso-Fernandez, J. Fierrez, and J. Ortega-Garcia are with Biometric Recognition Group – ATVS, Escuela Politecnica Superior, Universidad Autonoma de Madrid, 28049 Madrid, Spain (e-mail: fernando.alonso@uam.es; julian.fierrez@uam.es; javier.ortega@uam.es).

O. Ambekar is with CentrumWiskunde and Informatica (CWI), 1098 XG, Amsterdam, The Netherlands.

J. Baker and D. Maurer are with the Applied Physics Laboratory, Johns Hopkins University, Laurel, MD 20723 USA (e-mail: Donald.Maurer@jhuapl.edu; John.P.Baker@jhuapl.edu).

B. Dorizzi is with the Electronics and Physics Department, Institut Telecom, Telecom and Management SudParis, 91011 Evry, France (e-mail: Bernadette.dorizzi@it-sudparis.eu).

H. Ganster is with the Institute of Digital Image Processing, Joanneum Research, Graz 8010, Austria.

A. A. Salah is with the ISLA-ISIS, University of Amsterdam, 1098 XG Amsterdam, The Netherlands.

T. Scheida is with Applied/Media Informatics, IT Security, Department of Informatics and Media, Brandenburg University of Applied Sciences, D-14770 Brandenburg an der Havel, Germany.

C. Vielhauer is with Research Group Multimedia and Security, Department of Technical and Business Information Systems, Faculty of Computer Science, Otto-von-Guericke-University of Magdeburg, D-39106 Magdeburg, Germany (e-mail: claus.vielhauer@iti.cs.uni-magdeburg.de).

## I. Introduction

### A. Multimodal Biometric Systems

IN order to improve confidence in verifying the identity of individuals seeking access to physical or virtual locations, both government and commercial organizations are implementing more secure personal identification (ID) systems. Designing a highly secure and accurate personal identification system has always been a central goal in the security business. This challenge can be met by resorting to multimodal biometric systems [1]–[3] with the aim of increasing the security as well as identification performance. The multimodal biometrics aproach can be extended by explicitly considering the sample quality of the input biometric signals and weighting the various pieces of evidence based on objective measures of quality of the biometric traits. This formulation of the multimodal biometrics fusion problem is called quality-based fusion. It is a research topic of increasing importance.

*B. State-of-the-Art in Quality-Based Fusion*

The first known work on quality-based multimodal biometrics is [4], which presented the problem under a framework of Bayesian statistics. The result was an Expert Conciliation scheme including weighting factors not only for the relative accuracy of the experts but also for the confidence of the experts regarding the scores for particular input samples. The idea of relating sample confidence values to actual quality measures of the input biometric signals was also introduced in that work, but nevertheless not experimentally studied, under the same framework, until [5]. The first experimental study on quality-based fusion was limited to the use of a chimeric bimodal database (where biometric traits of different persons are combined to form a virtual identity), with the use of quality measures manually generated by a human expert. A follow-up work by the same researchers overcame the experimental limitation and provided a more realistic experimental setup to show the benefits of incorporating quality-based information in standard fusion approaches based on discriminative [6], and generative learning [7].

The concept of confidence of matching scores was considered in [8]. In that work, the authors demonstrated the merit of using measures of confidence in fusion. This research line was further developed in [9], where confidence measures based on the margin between impostor and client score distributions were developed.

Another research direction in quality-based fusion reported in the literature is based on clustering [10]. In this work, quality measures obtained directly from the input biometric signals were used to fuzzify the scores provided by the different component biometric systems. The authors demonstrated that fuzzy versions of k-means and vector quantization including the quality measures tended to outperform the standard nonfuzzy clustering methods. This work, to the best of our knowledge, is the first one reporting experimental results of quality-based fusion.

A more recent effort in quality-based fusion was reported in [11], where the authors developed a scheme based on polynomial functions. Quality measures were introduced in the optimization problem for training the polynomials as weights in the regularization term.

Other recent advances in quality-based fusion for multimodal biometrics are based on the following concepts: logistic regression with quality measures as features [12], Bayesian credence [13], Bayesian belief networks with quality measures as inputs [14], and joint score and quality classifiers using the likelihood ratio test [15], [16].

*C. Motivation*

The motivation for carrying out this study is as follows.

1) *The need for benchmarking quality-based fusion*: While there are quite a few papers on quality-based fusion, e.g., [4]–[6], [12], [13], and [16], to the best of our knowledge, there exists no benchmark database on which these algorithms can be compared and on which real progress can be measured. Note that although the existence of multimodal corpora is a necessary prerequisite of benchmarking multimodal and multialgorithmic (i.e., employing different algorithms on the same biometric data) fusion algorithms, it is not sufficient. For instance, it is not straightforward to compare two fusion algorithms in the case where each algorithm relies on its own set of biometric subsystems. This is because an observed improvement of a particular fusion algorithm may be due to the superior performance of its biometric subsystems rather than the merits of the fusion process itself. There is, therefore, a need for benchmarking fusion algorithms on a common ground, e.g., using the same biometric match scores for the score-level fusion, or some common features for the feature-level fusion.

2) *The cost implications*: While using more biometric devices and/or samples can increase the system performance, as demonstrated in [1], [17], [18] (and references therein) and elsewhere in the literature, such improvement often comes at the expense of acquiring more data, and, therefore, incurring more processing time, and adding the cost of hardware and its maintenance. All these aspects add up to much higher global operating costs. In this study, the abstract concept of cost is formally defined and is used to assess multimodal biometric fusion against such cost.

3) *System robustness*: We expect that using more biometric systems/devices can increase the robustness of the combined system against spurious verification errors (false acceptance and false rejection) of any single subsystem/device, e.g., [1] and references therein. In addition, such a combined multimodal system can also robustly handle software or operational failures such as failure-to-extract or failure-to-match, or even hardware failure (complete breakdown), resulting in invalid or missing match scores. The latter property has not been thoroughly investigated.

*D. Cost-Sensitive and Quality-Based Fusion Evaluation Campaign*

The above aspirations have been addressed by designing a benchmark database and organizing an evaluation campaign using the database. Seven teams participated in our evaluation, and altogether, they submitted 22 fusion algorithms.

The campaign was divided into two parts: quality-dependent evaluation and cost-sensitive evaluation. The first challenge was to evaluate quality-dependent fusion algorithms whereas the second involved evaluating conventional fusion algorithms. In both cases, we considered also the possibility of investigating two further subproblems. The first one involved client-specific or user-dependent fusion where one can train a fusion classifier that is tailored to each identity claim (see, e.g., [19] and [20] for a more comprehensive survey). The second was concerned with handling missing information. For instance, when one or more biometric subsystems are not operational due to failure-to-acquire or failure-to-match a biometric sample, we require the fusion system to be able to output a combined score. This is feasible because the subsystem match scores contain redundant information (i.e., each can be seen as a support for the same hypothesis although their actual values may be in different ranges).

An obvious disadvantage of score-level fusion is that, by using only scores, a lot of precious nonclass discriminatory information is lost, e.g., the quality of raw biometric signal. Here are two examples: a person's face can change drastically with illness, diet, or age, as well as with the application of cosmetics, a change in hair color or style, or a sun tan; and a person's voice can differ significantly with congestion caused by a cold. This information is *nonclass discriminatory* because it cannot be used to distinguish different individuals. Quality measures are expected to provide measurements designed to capture these changes in ways that could usefully be exploited in the fusion process. In this sense, quality measures should in theory compensate for the loss of information without sacrificing the practical advantage offered by score-level fusion. In practice, however, tapping the quality information, which is nonclass discriminatory in nature, in order to improve the classification performance, is not a trivial problem.

In a cost-sensitive evaluation scheme, one considers a fusion task as an optimization problem whose goal is to achieve the highest performance (as a function of false acceptance and false rejection decisions) at a desired minimum cost. We refer to "cost" as the price paid for acquiring and processing information, e.g., requesting samples from the same device or using more biometric devices (which entails longer processing time). In this situation, a decision can be made even if not all the subsystem match scores are observed. Therefore, the proposed cost-based evaluation effectively considers the redundancy of multimodal or multialgorithmic information. This subject has not been adequately addressed in the literature on multimodal biometrics such as [1], [17], and [18] because in the work, it is assumed that all the match scores are available.

### E. Contributions

The contribution of this paper is multifold.
1) *Benchmark for multimodal biometric quality-based fusion*: Thanks to the participation of seven teams, the BioSecure multimodal biometric evaluation campaign received 22 fusion systems for comparison. To the best of our knowledge, this is the first time such a comparison has been carried out.
2) *Cost-sensitive and robustness evaluation*: We formally introduce the problem of cost-sensitive evaluation of multimodal fusion algorithms. Normally, the common assumption behind multimodal biometric fusion is that all match scores are observed. However, in reality, as match scores and quality measurements are generated, failures to extract or to match can occur. Such failures are common especially when biometric templates and query samples are acquired using different devices.
3) *Sequential fusion*: This novel approach takes match scores into account sequentially until a desired level of confidence is reached, or until all the match scores are exhausted. The algorithm is ideally suited to minimize the cost of multimodal biometrics by dynamically controlling the number of matching performed.
4) *A benchmark database of multimodal biometric score and quality measures*: The data sets used to benchmark fusion algorithms have been made publicly available at http://face.ee.surrey.ac.uk/fusion. The most similar effort to our attempt is the XM2VTS score-level fusion benchmark data set [21]. However, the latter does not contain any quality measures for each biometric modality.

### F. Paper Organization

The paper is organized as follows: Section II categorizes and summarizes the submitted classifiers. The BioSecure DS2 data set (with score and quality measures) is detailed in Section III. Section IV explains the two evaluation protocols. The results of the evaluation can be found in Section V. This is followed by conclusions in Section VI.

## II. BRIEF DESCRIPTION OF SUBMITTED SYSTEMS

This evaluation involves 22 submissions of fusion systems from seven sites. We will begin by introducing a common notation and then describing the submitted fusion algorithms using this notation. A complete list of the systems is shown in Table I.

### A. Classifier Categorization and Notation

Let $y_i \in \mathbb{R}$ be the output of the $i$th biometric subsystem and let there be $N$ biometric subsystem outputs, i.e., $i \in \{1, \ldots, N\}$. For simplicity, we denote $\mathbf{y} = [y_i, \ldots, y_N]'$, where the symbol "$'$" is the matrix transpose operator. The most commonly used fusion classifier in the literature takes the following form:

$$f : \mathbf{y} \rightarrow y_{\text{com}} \qquad (1)$$

where $y_{\text{com}} \in \mathbb{R}$ is a combined score. We shall refer to this classifier throughout this report as a *conventional* fusion classifier.

If the function $f$ takes into consideration the signal quality, then $f$ is considered a *quality-dependent* fusion classifier. Let the signal quality of the $i$th biometric subsystem be represented by a vector of $L_i$ measurements, $\mathbf{q}_i \in \mathbb{R}^{L_i}$. Note that different biometric subsystems may have a different number of quality measures $L_i$. For simplicity, we denote $\mathbf{q}$ as a concatenation of all $\mathbf{q}_i$'s, i.e., $\mathbf{q} = [\mathbf{q}_1', \ldots, \mathbf{q}_N']'$. The function $f$ in this case takes the following form:

$$f : \mathbf{y}, \mathbf{q} \rightarrow y_{\text{com}}. \qquad (2)$$

Any fusion classifier can be categorized into one of the two forms just mentioned.

The function $f$ can be a *generative* or a *discriminative* classifier. In the former case, class-dependent densities are first estimated and decisions are taken using the Bayes rule or the Dempster–Shafer theory. In the latter, the decision boundary is directly estimated. A common characteristic of both types of classifiers is that the dependency among observations (scores or quality measures) is considered.

There exists also another approach that we will refer to as the *transformation-based approach* [22] which constructs a fusion classifier in two stages. In the first stage, the match scores of each biometric subsystem are independently transformed into a comparable range, e.g., in the range [0, 1]. In the second stage, the resulting normalized scores of all biometric subsystems are combined using a fixed rule such as sum or product [23]. The transformation-based fusion classifier based on the sum rule, for instance, has the following form:

TABLE I
LIST OF SUBMITTED SYSTEMS

| System name | Q | C | Characteristics |
|---|---|---|---|
| AMSL-BIO InvLW (Ref. Sec II-B.5) | × | √ | transformation based, weighted sum rule is used with weights being inversely proportional to Equal Error Rate empirically calculated from the development set. See [34]. |
| AMSL-BIO QW (Ref. Sec II-B.5) | √ | × | transformation based, weighted sum rule is used with weights being inversely proportional to the quadratic term of Equal Error Rate empirically calculated from the development set. See [34]. |
| CWI SVM (Ref. [35]) | √ | √ | discriminative classifier, quality-independent, train with high quality data only |
| CWI IMOFA (Ref. Sec II-B.1) | √ | √ | Bayesian classifier (generative) whose class-conditional densities are each independently estimated using an Incremental Mixture of Factor Analyzers [26]. The classifier estimates the posterior probability of client given the observed scores. It is quality-independent and uses sequential strategy with double thresholds (for cost-sensitive evaluation). |
| JHUAPL (Ref. Sec II-B.1) | √ | × | Bayesian belief network (generative classifier), quality-dependent, use all available scores. See [14] |
| JR (Ref. Sec II-B.3) | √ | √ | Dempster-Shafer theory of evidence. A biometric subsystem is chosen to participate in the fusion if it maximizes the following criterion: $$benefit = y_i * (2 - cost). \qquad (24)$$ where cost takes on a value of one for a different device or 0.3 for reusing the same device. More explanation about the cost assignment can be found in Section IV-A. |
| GET 1 (Ref. Sec II-B.1) | √ | √ | Bayesian classifier with equal class priors was used, i.e., (5) with $P(\texttt{C}) = P(\texttt{I}) = 0.5$. The density $p(y\|k)$ is estimated using a mixture of Gaussian components [25]. In case of missing values, scores are independently normalized into $[0, 1]$ using the empirically observed minimum and maximum value, known as the Min-Max normalization [22]. The average rule is used to combine the normalized scores. See [37] |
| GET 2 (cost) (Ref. Sec II-B.1) | × | √ | Similar to GET 1 (cost) except that scores are taken into account sequentially until all channels of data are exhausted or the combined score is smaller than lower threshold or greater than a higher threshold. See [38]. |
| GET 2 (quality) | √ | × | Similar to GET 1 except that each fusion classifier is designed specifically for each of the four cases as mentioned in Section IV-B. |
| GET 3 (quality) (Ref. Sec II-B.1) | √ | × | This is an enhanced version of GET 2 (quality) which also employs sequential selection of channels (as in the GET 2 cost-based system). |
| UniS qfixed (Ref. Sec II-B.6) | √ | √ | transformation-based, quality-dependent, selectively switches between sum and product rule depending on the quality measures. See [39]. |
| UniS qfused (Ref. Sec II-B.4) | √ | √ | logistic regression (discriminative) modeling the posterior probability of being a client given scores and quality measures as observation, i.e., $P(C\|y, q)$. See [12]. |
| UniS Naive Bayes (Ref. Sec II-B.4) | × | √ | Discriminative classifier, modeling $\prod_i P(C\|y_i)$ where $P(C\|y_i)$ is estimated using logistic regression. See [28] for Naive Bayes and [40] for logistic regression. |
| UniS BNQ (Ref. Sec II-B.2) | × | √ | Generative classifier that estimates the device identity directly (see (10)). The sum rule is used to combine all $y_i^{norm}$'s that are observed. See [30]. |
| UPM (cost) (Ref. Sec II-B.4) | × | √ | transformation based, with each transformation function (one for each channel of data) being a logistic regression, mapping scores into log-likelihood ratios. The max rule is used to combine the scores. See [33]. |
| UPM (quality) (Ref. Sec II-B.4) | √ | × | Similar to its cost-based counterpart, except that the transformation function is device-specific for the face and device-independent for the fingerprint. The max rule is used to combine the scores. See [33]. |

Legend: Q = Quality-based evaluation; C = cost-sensitive evaluation
√ means used in quality-based and/or cost-sensitive evaluation; × means not used
AMSL-BIO is a submission from University of Magdeburg; GET from Telecom and Management SudParis (formerly GET - Institut National des Telecommunications); UPM from Universidad Politecnica de Madrid; UniS from University of Surrey; JR from Joanneum Research; CWI from Centrum voor Wiskunde en Informatica; and JHUAPL from Johns Hopkins University Applied Physics Laboratory.

$$y_{\text{com}} = \sum_i f_i^{\text{norm}}(y_i) \qquad (3)$$

where $f_i^{\text{norm}}$ is the transformation function of the $i$th biometric subsystem output. The one using the product rule can be realized by replacing the sum ($\sum_i$) in (3) with a product ($\prod_i$). For product fusion, the normalized scores have to satisfy certain properties (i.e., being non-negative and in a given range). The same applies to the minimum ($\min_i$), the maximum ($\max_i$), and the median rules, i.e., taking the minimum, the maximum, and the median of the transformed values, respectively. These approaches have been investigated in [22]. Among the fixed rules, the sum rule is commonly found to be the most effective in many applications. By interpreting the classifier outputs as probabilities, Kittler et al. [23] showed that the sum rule is resilient to errors affecting the individual classifier outputs. An intuitive explanation is that by summing two variables (regardless of whether the output is in probability or not), the resultant summed variable will have smaller variance than the average variance of the individual variables. The extent of the reduction of variance of the combined variable (fusion output) depends on the dependency (correlation) among the variables (classifier outputs) contributing to the sum. Using the average rule (which defers from sum by a negligible constant factor), the relationship between the correlation among the constituent classifier outputs and the fusion classifier performance was clarified by Poh and Bengio in [24].

In each of the above three cases, the following decision function can be used:

$$\text{decision}(\mathbf{y}) = \begin{cases} \text{accept}, & \text{if } y_{\text{com}} > \Delta \\ \text{reject}, & \text{otherwise} \end{cases} \qquad (4)$$

where $\Delta$ is a global decision threshold. We replace $\Delta$ with $\Delta_j$ for the $j$th user if the decision threshold is client-specific.

In some of our discussions, it will be convenient *not to* distinguish between the conventional and quality dependent fusion classifiers. This will be achieved by using common notation. In particular, for the conventional classifier, this can be further described by the feature vector $\mathbf{x}_i$, which is defined as $\mathbf{x}_i \equiv \mathbf{y}_i$, where $\mathbf{y}_i$ is a vector of biometric subsystem outputs dependent on the $i$th biometric modality. The feature vector of a quality-dependent fusion classifier, on the other hand, will take the input $\mathbf{x}_i \equiv [\mathbf{y}_i', \mathbf{q}_i']'$ instead. Following the same convention as before, we will write $\mathbf{x}$ as a concatenation of all $\mathbf{x}_i$'s, i.e., $\mathbf{x} = [\mathbf{x}_1', \ldots, \mathbf{x}_N']'$.

In the discussion that follows, we will elaborate several forms of $f$ used in our evaluation. We will, however, not discuss the algorithms in great detail but instead we shall attempt to capture the key intuitive ideas behind the methods. For details, the interested reader should refer to the relevant citations.

### B. Submitted Systems

*1) Generative Classifier Estimating the Posterior Probability:* For the generative classifier, one can infer the probability (or posterior probability) of being a client using the following Bayes rule:

$$y_{\text{com}} = P(\mathtt{C}|\mathbf{x}) = \frac{p(\mathbf{x}|\mathtt{C})P(\mathtt{C})}{p(\mathbf{x}|\mathtt{C})P(\mathtt{C}) + p(\mathbf{x}|\mathtt{I})P(\mathtt{I})} \quad (5)$$

where $p(\mathbf{x}|k)$ is the density of $\mathbf{x}$, or a likelihood function conditioned on the class label $k$ which is either client or impostor, i.e., $k \in \{\mathtt{C}, \mathtt{I}\}$, and $P(k)$ is the prior class probability. Being a probability, $y_{\text{com}}$ is in the range $[0, 1]$. This classifier is also known as a Bayes classifier. The Bayes optimal decision threshold as defined in (4) is $\Delta = 0.5$.

Referring to Table I, the fusion classifier GET-1 and CWI-IMOFA are of this form. In GET-1, the likelihood function $p(\mathbf{x}|k)$ was estimated using a mixture of Gaussian components, or a Gaussian mixture model (GMM) [25], i.e.,

$$p(\mathbf{x}|k) = \sum_{c=1}^{N_{\text{cmp}}^k} w_c^k \mathcal{N}\left(\mathbf{x}|\boldsymbol{\mu}_c^k, \boldsymbol{\Sigma}_c^k\right) \quad (6)$$

where the $c$th component class-conditional (denoted by $k$) mean vector is $\boldsymbol{\mu}_c^k$ and its covariance matrix is $\boldsymbol{\Sigma}_c^k$. There are $N_{\text{cmp}}^k$ components for each $k = \{\mathtt{C}, \mathtt{I}\}$.

In CWI-IMOFA, a mixture of factor analyzers [26] was used instead. While the density estimated can also be written as in (6), the number of parameters needed is effectively smaller. This is because a factor analyzer assumes that a small number of low-dimensional latent variables (factors) $\mathbf{z}$ cause the correlation that is gauged by the $c$th component conditioned on class $k$. Dropping the superscript $k$ for notational economy (since each term is conditioned on the class $k$), each factor analyzer component can be described by

$$\mathbf{x} - \boldsymbol{\mu}_c = \boldsymbol{\Lambda}_c \mathbf{z} + \boldsymbol{\epsilon}_c. \quad (7)$$

$\boldsymbol{\Lambda}_c$ is called the *factor loading matrix* (for component $c$). This matrix characterizes the dependency of data points on each factor. $\boldsymbol{\epsilon}_c$ is the Gaussian noise and is assumed to be distributed $\mathcal{N}(0, \Psi)$, where $\Psi$ is a diagonal matrix, interpreted as sensor noise common to all components. When $\mathbf{x}$ is $d$-dimensional, $\boldsymbol{\Sigma}_c$ is $d \times d$, whereas with $p < d$ factors, $\boldsymbol{\Lambda}_c$ is $d \times p$.

If there are missing observations in $\mathbf{x}$, one can still calculate the marginal distribution of the observed features by marginalizing away the missing features. For a Gaussian mixture distribution, calculating its marginal distribution can be achieved efficiently by manipulating the Gaussian mean and covariance matrices [27], thereby, dispensing with the need for explicit integration. Such an approach is implemented by the organizer (UniS), referred to as the GMM-Bayes classifier.

In contrast to UniS's GMM-Bayes classifier, the GET-1, 2, and 3 submissions did not deal with the missing observation using Gaussian marginals. Instead, whenever there is a missing observation, GET's GMM-Bayes systems compute the fused score using a transformation-based strategy. First, each system output is normalized to the range of $[0, 1]$ using the empirically observed minimum and maximum value (Min–Max normalization) [22]. Then, the average rule is used to combine the normalized scores. Such a strategy works because the outputs of the transformation-based fusion approach and that of the GMM-Bayes classifier are in the same range, i.e., $[0, 1]$, although only the latter case can be interpreted as probability.

The CWI-IMOFA submission dealt with the missing values by replacing them with their corresponding median values. The authors found experimentally (on the development set) that this did not affect the generalization performance significantly.

*2) Generative Classifier Using the Log-Likelihood Ratio Test:* An alternative generative approach based on the log-likelihood ratio test, which relies on the Neyman–Pearson lemma [28], takes the following form:

$$y_{\text{com}} = \log \frac{p(\mathbf{x}|\mathtt{C})}{p(\mathbf{x}|\mathtt{I})}. \quad (8)$$

Its associated decision threshold [as in (4)] is optimal when

$$\Delta = -\log \frac{P(\mathtt{C})}{P(\mathtt{I})}.$$

In practice, the output $y_{\text{com}}$ is a real (positive or negative) number in the range of hundreds or thousands.

Adopting the Naive Bayes strategy, (8) can be computed as

$$y_{\text{com}} = \log \frac{\prod_i p(\mathbf{x}_i|\mathtt{C})}{\prod_i p(\mathbf{x}_i|\mathtt{I})} = \sum_i \log \frac{p(\mathbf{x}_i|\mathtt{C})}{p(\mathbf{x}_i|\mathtt{I})} = \sum_i y_i^{llr} \quad (9)$$

where we defined

$$y_i^{llr} \equiv \log \frac{p(\mathbf{x}_i|\mathtt{C})}{p(\mathbf{x}_i|\mathtt{I})}.$$

We shall now deal with two cases: conventional fusion where $\mathbf{x}_i = y_i$, and quality-based fusion where $\mathbf{x}_i = [\mathbf{q}_i', y_i]'$. In the first case, the function is $y_i \to y_i^{llr}$. This is a one-to-one mapping function (for each modality), therefore, a possible function for $f_i^{\text{norm}}: y_i \to y_i^{llr}$ [see (3)]. This can be seen as a transformation-based approach with the fusion operator being the sum rule. Such a Naive Bayes classifier was provided by UniS.

The second (quality-dependent) case has been reported in the literature [29]. Note that the observation vector $\mathbf{x}_i = [\mathbf{q}_i', y_i]'$ has $1+\mathtt{L}_i$ dimensions. This increased dimensionality, especially in the case $\mathtt{L}_i \gg 1$ (due to $\mathbf{q}_i$), can possibly pose a potential estimation problem, i.e., modeling the increased number of dimensions may be less effective as it is usually not supported by the required exponential increase in the number of training samples (in the worst case). In fact, there is only a fixed number of training samples to design a fusion classifier.

Apart from the baseline Naive Bayes fusion, the UniS submission also includes a version that considers the quality measures in its density estimation but does not suffer from the above-mentioned increased dimensionality of the $[\mathbf{q}'_i, y_i]'$ space. The model used here assumes that the biometric subsystem outputs and the quality measures are *conditionally* independent given the quality state, represented by $Q$. The authors defined a quality state to be a *cluster* of quality measures that are "similar" to each other [16]. The rationale of this model is that data belonging to the same cluster of quality measures will share the same characteristic. Therefore, it is sensible to construct a fusion strategy for each cluster of data (of similar quality). The consequence of this is that the complexity of the fusion classifier now is no longer *directly dependent* on the dimension of the quality measures $\mathbf{q}_i$, but is dependent on the number of clusters. In their submitted implementation, these clusters are found using a GMM model (6), but constrained to be between one and three (as a means to control the model complexity). The GMM parameters are estimated using the expectation maximization (EM) algorithm. For each of the biometric subsystem outputs $y_i$, the following quality-based normalization output is computed:

$$y_i^{llr} = \log \frac{\sum_Q p(y_i|\texttt{C},Q)P(Q|\mathbf{q}_i)}{\sum_Q p(y_i|\texttt{I},Q)P(Q|\mathbf{q}_i)} \quad (10)$$

where $Q$ denotes a cluster of quality measures. The sum over all the quality states of $Q$ is necessary since the quality state is a hidden variable; only $y_i$ and $\mathbf{q}_i$ are observed. Note that the dimensionality involved in estimating $p(y_i|k,Q)$ is effectively one since $y_i$ is one-dimensional and $Q$ is a discrete variable. The partitioning function $P(Q|\mathbf{q}_i) : \mathsf{L}_i \to \mathbb{R}$ refers to the posterior probability that $\mathbf{q}_i$ belongs to the cluster $Q$. This term is also known as *responsibility* in the GMM literature [25].

There are at least three possible ways of clustering the quality measures:

- In the first approach, one divides the quality measures according to the device which was used to collect the biometric samples (from which the quality measures have been derived). If there are $N_d$ devices, there will be $N_d$ clusters of quality measures. In this case, we say that the clusters $Q$ are device-dependent. This approach was first reported in [30].
- The second approach can be considered a further refinement of the first approach. Since there is no guarantee that the quality is consistent over all devices, it may be desirable to further find the natural clustering that exists for each device-dependent cluster of quality measures. For the case of a single device, such a study was reported in [16] where in order to combine several face experts (hence intramodal fusion), a separate fusion strategy was devised for each cluster of quality measures.
- The third approach would consider finding a natural clustering of quality measures in a device independent manner.

The UniS submission in Table I is based on the first approach. Since the device is known during training (but not during testing), instead of using EM to infer the quality state, the quality state is probabilistically estimated, i.e., $P(Q|\mathbf{q}_i)$ is estimated using a supervised approach. In the implementation of (10), the following Bayesian classifier (with equal class priors) was used:

$$P(Q|\mathbf{q}_i) = \frac{p(\mathbf{q}_i|Q)}{\sum_{Q_*} p(\mathbf{q}_i|Q_*)}. \quad (11)$$

where the device-specific density of quality measures, $p(\mathbf{q}_i|Q_*)$ (for a given $Q_*$), was modeled using a GMM.

Note that the three procedures mentioned above, i.e., (9) (including both the conventional and quality-based fusion) and (10), output $y_i^{llr}$ for $i = 1, \ldots, N$ that are directly combined using the sum rule (9). Hence, if a particular subsystem fails to output any score observation, the corresponding terms in the sum will be missing, hence not contributing to the final output.

The submission by JHUAPL can be interpreted as a realization of (9) (except that log was not used) and is very similar to (10) which involves estimating the density $p(y_i|k,Q)$. Instead of clustering $Q$ using EM, the authors binned the data to compute a histogram. The binning process can only work for *scalar* quality measures but not for a *vector* of quality measures. In order to generalize to the latter case, one must resort to the clustering approach.

*3) Generative Classifier Using the Dempster–Shafer Theory:* The Dempster–Shafer theory of evidence [31] attempts to reflect uncertainty of information sources by degrees of belief. The information sources, or frames of discernment, in our case, refer to the class labels $k = \{\texttt{C}, \texttt{I}\}$. While in the Bayesian theory, these two events are disjoint, in the Dempster–Shafer theory, one considers all possible combinations, i.e., $2^{\{\texttt{C},\texttt{I}\}} = \Psi = \{\{\texttt{C},\texttt{I}\}, \{\texttt{C}\}, \{\texttt{I}\}, \phi\}$, noting that information sources, unlike events in the usual probability interpretation, are not necessarily disjoint. However, similar to probability, only one unit of mass is distributed among all the possible information sources. Let $m : A_i \to \mathbb{R}$ be the function that assigns a mass to the information source $A_i$. The function $m$ is subject to the following constraints:

$$\begin{aligned} 0 &\leq m(A_i) \leq 1, \quad A_i \in \Psi \\ m(\phi) &= 0 \\ \sum_{A_i \in \Psi} m(A_i) &= 1. \end{aligned} \quad (12)$$

A new mass assignment $m$ is then combined with the mass distribution $m_{\text{prev}}$ derived from all previous information sources using *Dempster's Rule of Combination*, i.e., (13). The result of this task is a new distribution, $m^*$, that incorporates the joint information provided by the sources selected up to this moment

$$\begin{aligned} m^*(A_i) &= (m \oplus m_{\text{prev}})(A_i) \\ &= \frac{1}{1 - \text{Conflict}} \sum_{A_p \cap A_q = A_i} m(A_p) m_{\text{prev}}(A_q) \end{aligned} \quad (13)$$

where

$$\text{Conflict} = \sum_{A_p \cap A_q = \phi} m(A_p) m_{\text{prev}}(A_q). \quad (14)$$

Note that the sum in (13) iterates over all the information sources containing $A_i$. The conflict term is a normalizing factor and displays the conflict between the new evidence and the actual knowledge.

After selecting one information source (i.e., one biometric score value), the mass distribution is derived in the following way. The score value is interpreted as a percentage of certainty that the claimed identity is true, and thus assigned to the set C (client or genuine user). Another portion of mass is assigned to the set containing the whole frame of discernment (15)

$$m(\{\text{C}\}) = y_i^{\text{norm}}$$
$$m(\{\text{C}, \text{I}\}) = 1 - y_i^{\text{norm}} \quad (15)$$

noting that $y^i$ are normalized into the range $[0, 1]$. The combined scores can be expressed by

$$y_{\text{com}} = m_{\text{com}}(\{\text{C}\}) \text{ with } m_{\text{com}} = \bigoplus_i m_i \quad (16)$$

where $m_i$ is the mass distribution defined according to (15), and $\bigoplus_i$ denotes the application of Dempster's rule of combination (13) for all selected biometric subsystem outputs. Note that $m_i$ with missing observation will not participate in the Dempster's rule of combination. This operation is analogous to the sum rule in probability, i.e., (9).

This submission was provided by JR (see Table I).

*4) Discriminative Classifier Using Linear Logistic Regression:* LR is defined as:

$$y_{\text{com}} \equiv P(\text{C}|\mathbf{x}) = \frac{1}{1 + \exp(-g(\mathbf{x}))} \quad (17)$$

where

$$g(\mathbf{x}) = \sum_{j=1}^{M} \beta_j x_j + \beta_0 \quad (18)$$

where $x_j$'s are elements in $\mathbf{x}$. The weight parameters $\beta_j$ are optimized using gradient ascent to maximize the likelihood of the training data given the LR model [32]. It can be shown that the following relationship is true:

$$g(\mathbf{x}) = \log \frac{P(\text{C}|\mathbf{x})}{P(\text{I}|\mathbf{x})}. \quad (19)$$

We shall introduce the Naive Bayes version of logistic regression. This can be done by replacing $\mathbf{x}$ in (18) with $\mathbf{x}_i$, recalling that $\mathbf{x}_i$ represents the modality-dependent observation (which can be score alone, or score augmented with the quality measures derived from the same modality). By using the Naive Bayes assumption, one can combine $g_i(\mathbf{x}_i)$ for $\mathbf{x}_i$ of different modality $i$ in the following way:

$$y_{\text{com}} = \sum_i y_i^{llr} = \sum_i g_i(\mathbf{x}_i) = \sum_i \log \frac{P(\text{C}|\mathbf{x}_i)}{P(\text{I}|\mathbf{x}_i)}. \quad (20)$$

The submission from UPM used a version of LR in a way that incorporates the quality measures [33]. The fundamental idea is to design a device-dependent score-normalization strategy via LR for each device. During inference, the acquisition device is inferred so that the score is normalized according to the inferred device. The resulting normalized match scores are combined using (20). In this context, the term $\mathbf{x}_i$ in (20) contains only the score $y_i$ and the quality measurement $\mathbf{q}_i$ is not used.

The device-dependent normalization procedure is defined by

$$y_i^{llr} = g_{Q_*}(y_i) \quad (21)$$

where $g_Q$ is the function shown in (18) except that one such function is created for each cluster of *device-dependent* quality $Q$. Among all the possible states of $Q$ (one such $Q$ being associated with a device), the chosen $Q_*$ is selected so as to maximize $P(Q|\mathbf{q}_i)$, the posterior probability of $Q$ given the quality vector $\mathbf{q}_i$

$$Q_* = \arg\max_Q P(Q|\mathbf{q}_i).$$

The "UniS qfuse" submission [12] is based on the logistic regression shown in (17) except that the vector $\mathbf{x}$ is a reduced second-order polynomial expansion between the score vector $\mathbf{y}$ and the quality measures $\mathbf{q}$, i.e., $\mathbf{x}_i \equiv [y_i, \mathbf{q}_i', (y_i \otimes \mathbf{q}_i)']'$, where $\otimes$ is known as a *tensor product* whose output is a single column vector. According to this operator, each element in $y_i$ is multiplied with each element in $\mathbf{q}_i$. Note that the proposed method does not take into consideration $[y_i \times y_i]'$ nor $[\mathbf{q}_i \otimes \mathbf{q}_i]'$ in order to keep the number of parameters to be estimated small, while at the same time allowing the model to gauge the interaction between scores and quality measures. The final input to the logistic regression is $\mathbf{x} = [\mathbf{x}_1', \ldots, \mathbf{x}_N']'$. In so doing, one also avoids the need to model the unnecessary interaction among elements in $y_i$ and $\mathbf{q}_j$, i.e., $[y_i \otimes \mathbf{q}_j]'$, where $i \neq j$ are indexes of different subsystem outputs.

*5) Error-Based Fusion Classifier:* The submission by AMSL-BIO InvLW attempts to combine subsystem outputs based on their authentication performance on the development set. The fusion score is constructed as a linear combination of matching scores, i.e.,

$$f_{\text{com}} = \sum_i f_i^{\text{norm}}(y_i) w_i$$

where $f_i^{\text{norm}} :\to [0, 1]$ is known as a score normalization procedure and $w_i$ is the weight associated to the $i$th subsystem subject to the constraint $\sum_i w_i = 1$.

The following linear weight is used for each $i$th subsystem [34]

$$u_i = \begin{cases} \frac{\text{eer}_i}{\sum_{j=1}^{N} \text{eer}_j}, & \text{if } \frac{\text{eer}_i}{\sum_{j=1}^{N} \text{eer}_j} \geq \frac{0.5}{N} \\ \infty, & \text{otherwise} \end{cases} \quad (22)$$

where $w_i = 1/u_i$ and $\text{eer}_i$ is the equal error rate (EER) of the $i$th subsystem measured on the development score set. The first condition ensures the subsystem whose weight is smaller than $0.5/N$ (recalling that $N$ is the number of subsystems) will not contribute to the final score because its output is considered insufficiently useful. In order to deal with missing matching scores, mean values of impostor and genuine score distributions are calculated. A missed score is replaced with the average of these two values.

*6) UniS: Fixed Rule Quality Dependent Fusion Classifier:* The subsystem outputs **y** are divided into two groups: those of higher quality and those of lower quality. Let us denote these two groups by $\{y_p^{\text{high}}\}$ and $\{y_q^{\text{low}}\}$, where $p \in \{1, \ldots, P\}$ and $q \in \{1, \ldots, Q\}$ are indexes of the subsystem outputs. The idea is that one combines the groups of similar quality (high or low) using the sum rule whereas among the groups of different scores using the product rule. The justification is that scores of different quality tend to disagree, implying higher independence and so the product rule may be more effective in this case. Similarly, scores of similar quality (high or low) tend to agree with each other, implying higher dependence and so the sum rule may be more effective. The resulting combined scores can be written as:

$$y_{\text{com}} = \begin{cases} \frac{1}{P} \sum_m y_p^{\text{high}} \times \frac{1}{Q} \sum_n y_q^{\text{low}}, & \text{if } P > 0 \text{ and } Q > 0 \\ \frac{1}{P} \sum_m y_p^{\text{high}}, & \text{if } Q = 0 \\ \frac{1}{Q} \sum_n y_q^{\text{low}}, & \text{if } P = 0. \end{cases} \quad (23)$$

The last two cases take care of the situation where one group or the other is not observed. A score is considered of high quality if its corresponding quality measure is higher than $\overline{q_i} - \sigma_{q_i}$, where $q_i \in \mathbb{R}$ is a quality measure, $\overline{q_i}$ is the average of $q_i$ and $\sigma_{q_i}$ is its standard deviation.

In case of an array of quality measures, the quality measures are normalized using the Min–Max normalization, and the average is computed to represent the quality measure for a sample.

*7) Other Classifiers:* The list of classifiers covered here is not meant to be exhaustive, but considers only the submitted systems. Another submitted fusion classifier is CWI-SVM's support vector machine (SVM), a discussion of which can be found in [35]. SVM has also been used elsewhere in the literature for biometric fusion, e.g., [36]. Finally, a general survey of fusion classifiers can be found in [1].

## C. Sequential Fusion

Sequential fusion is a fusion strategy where the match scores to be combined are considered *sequentially*, until a certain level of confidence is reached, or all the match scores are exhausted. The motivation is that for most access attempts (samples), a very few match scores are needed to reach a high level of confidence. In contrast, samples that are classified with low confidence are those that are found near the decision boundary and are necessarily sparse. Since significantly fewer number of match scores are needed, on average, for each access attempt, the overall cost is expected to be lower.

Among the submitted systems, only CWI-IMOFA and GET-2 (cost) adopt sequential fusion strategies. In both systems, the training of the system remains the same. However, during testing, different strategies are used. Both rely on an upper and a lower threshold, corresponding to the desired levels of confidence to accept a client and to reject an impostor, respectively.

For the CWI-IMOFA method, the modalities are considered in the following order: face, iris, fingerprint. The following decision strategies are adopted:
- If the face or the iris is present, do not take the fingerprints into account at all.
- If the posterior probability of observing a genuine user is higher than the upper threshold, at any time, output that probability and do not consider the rest of the modalities.
- If the posterior probability of observing a genuine user is lower than a percentage of the threshold (5% in the submitted version), output that probability and do not consider the rest of the modalities.

For the GET-2 systems, the following strategy was adopted. The face modality is used first. If a decision cannot be made, iris or fingerprint is then used depending on the face score and quality measures. Then, if a decision is not possible with the two match scores (and their associated quality measures), a third score is used and so on until the eight scores (and their associated quality measures) are exhausted [37].

## III. BIOSECURE DS2 DATA SET AND REFERENCE SYSTEMS

### A. BioSecure Database as a Test Bed

In order to realize the cost-sensitive and quality-based fusion evaluations mentioned in Section I-D, we constructed a database with scores as well as quality measures for each access using the BioSecure multimodal biometric database.[1] Since the input to all fusion algorithms is the same, they can be compared on equal grounds. To date, there exists no similar test bed suitable for *quality-dependent fusion algorithms* nor *cost-sensitive* evaluation. A similar work in this direction is [21]. However, the match scores were not supplemented by quality measures.

The BioSecure database was collected with the aim to integrate multidisciplinary research efforts in biometric-based identity authentication. Application examples of such an investigation are to meet the trust and security requirements for building access systems using a desktop-based or a mobile-based platform, as well as applications over the Internet such as teleworking and Web or remote-banking services. As far as data collection is concerned, three scenarios have been identified, each simulating the use of biometrics in remote-access authentication via the Internet (termed the "Internet" scenario), physical access control (the "desktop" scenario), and authentication via mobile devices (the "mobile" scenario). While the desktop scenario is used here, the proposed two evaluation schemes can equally be applied to the remaining two data sets.

The desktop scenario data set contains the following biometric modalities: signature, face, audio–video (PINs, digits, phrases), still face, iris, hand, and fingerprint. However, only still face, iris, and fingerprint are used for the evaluation schemes proposed here. This data set is collected from six European sites (only four are being used at the writing of this report). Although the data acquisition process is supervised, the level of supervision is extremely different from site to site.

This database contains two sessions of data separated by about a one-month interval. In each session, for each subject, two biometric samples are acquired per modality per device, hence resulting in four samples per modality per device (and per person) over the two sessions. There are several devices for the same biometric modality. The forgery data collected simulate PIN-reply attacks and imitation of dynamic signature (with several minutes of practice and with the knowledge of the signature dynamics). The volunteers are selected to have

---
[1]Available: http://biometrics.it-sudparis.eu/BMEC2007

TABLE II
LIST OF 24 CHANNELS OF MATCH SCORES) GENERATED FROM THE BIOSECURE DS2 DATABASE

| Label | template ID {n} | Modality | Template Sensor | Query Sensor | Remarks |
|---|---|---|---|---|---|
| fa | 1 | Still Face | web cam | web cam | Frontal face images (low resolution) |
| fnf | 1 | Still Face | CANON | CANON | Frontal face images without flash (high resolution) |
| fwf | 1 | Still Face | CANON | CANON | Frontal face images with flash (high resolution) |
| ir | 1–2 | Iris image | LG | LG | 1 is left eye; 2 is right eye |
| fo | 1–6 | Fingerprint | Optical | Optical | 1/4 is right/left thumb; 2/5 is right/left index; 3/6 is right/left middle finger |
| ft | 1–6 | Fingerprint | Thermal | Thermal | 1/4 is right/left thumb; 2/5 is right/left index; 3/6 is right/left middle finger |
| xfa | 1 | Still Face | CANON | web cam | Cross-device matching |
| xft | 1–6 | Fingerprint | Optical | Thermal | Cross-device matching. 1/4 is right/left thumb; 2/5 is right/left index; 3/6 is right/left middle finger |

For example, fo2 refers to the channel of data obtained by acquiring right index fingerprints using an optical fingerprint sensor. The web cam model is Phillips SPC 900. The model of CANON digital camera is EOS 30D. The iris capturing device is LG3000. The thermal sensor acquires a fingerprint as a subject sweeps his/her finger over it. The optical sensor acquires a fingerprint impression by direct contact (no movement required). The first seven rows show same-device matching and the last two show cross-device matching.

both genders in somewhat equal proportions of ages with the following distribution: 2/3 in the range 18–40 of age and 1/3 above 40.

Table II presents the 24 channels of data available. A *channel* of data is composed of the following quadruples: biometric trait, the acquisition device of a template, the acquisition device of a query sample, and matching algorithm. When the acquisition device used to prepare a template (during enrollment) is different from the one used to acquire a query sample, the matching is called *cross-device* matching, and the opposite is called *same-device* matching. There are 17 channels of data available under the same-device matching. For example, a left index fingerprint acquired using an optical fingerprint sensor is considered a channel of data. Using the notation presented in Table II, this channel of data is referred to as "fo5." The 17 channels of data, presented in order are $fa1, fnf1, fwf1, ir1, ir2, fo1, fo2, fo3, fo4, fo5, fo6, ft1, ft2, ft3, ft4, ft5$, and $ft6$. These 17 channels of data are all that is available to perform matching of data acquired using the same biometric device. The cross-device matching can only be performed on the face biometric and the six fingerprints (hence seven channels of data) because for each of these channels of data, two devices were available. These channels of data are prefixed with "x" (for "cross-device"). We only considered the scenario where the template data is acquired with a high-quality device whereas the query data is acquired with one of lower quality (post-determined by judging from the verification performance of the devices).

While there are 24 channels, we need only three reference systems, corresponding to the three chosen biometric modalities, i.e., face, fingerprint, and iris. We also need three pieces of software to extract their respective quality measures directly from the acquired images. Table III lists the reference systems of the three biometric modalities as well as their respective quality measures.

Among the 14 quality measures extracted, six are face-related quality measures (hence relying on a face detector), i.e., face detection reliability, spatial resolution between the eyes, presence of glasses, rotation in plane, rotation in depth and degree to which the face presentation is frontal. The remaining eight measures are general purpose image quality measures as defined by the MPEG standards. These quality measures were obtained using the Omniperception proprietary Affinity SDK.

TABLE III
REFERENCE SYSTEMS AND QUALITY MEASURES ASSOCIATED WITH EACH TO BIOMETRIC MODALITY

| Modality | Reference systems | Quality measures |
|---|---|---|
| Still Face | Omniperception's Affinity SDK[2] face detector; LDA-based face verifier | face detection reliability, brightness, contrast, focus, bits per pixel, spatial resolution (between eyes), illumination, background uniformity, background brightness, specular reflection, glasses, rotation in plane, rotation in depth and deviation from the frontal pose (all available from Omniperception's Affinity SDK) |
| Fingerprint | NIST Fingerprint system | texture richness [41] (based on local gradient) |
| Iris | A variant of Libor Masek's iris system | texture richness [42], difference between iris and pupil diameters and proportion of iris used for matching |

There is only a single fingerprint quality measure and it is based on the implementation found in [41]. It is an average fingerprint gradient computed over local image patches. When too much pressure is applied during fingerprint acquisition, the resulting fingerprint image usually has low contrast. Consequently, a minutia-based fingerprint matcher (which is the case for the NIST fingerprint system) is likely to under-perform with this type of image. The converse is also true for high contrast and clear fingerprint images.

Three iris quality measures are used. The first one, i.e., texture richness measure, is obtained by a weighted sum of the magnitudes of Mexican hat wavelet coefficients as implemented in [42]. The other two quality measures are functions of estimated iris and pupil circles. The first one is the difference between iris diameter and pupil diameter. If this difference is small, the iris area to be matched will be small, hence implying that the match scores may not be reliable. The second measure is the relative proportion of the mask diminishing the usable iris area for matching. A mask is needed to prevent matching on areas containing eyelashes and specular lights, for instance. Unfortunately, due to bad iris segmentation, and possibly suboptimal threshold to distinguish eyelashes from iris, our iris system is far from the performance claimed for Daugman's implementation [43].

## IV. EVALUATION PROTOCOL

The objective of an evaluation protocol is to provide a means to partition the development (or training) and the evaluation (test) data sets in such a way that fusion algorithms can be compared on equal grounds, avoiding optimistic performance bias of the test result.

The current release of the desktop scenario contains 333 persons. A newer version, yet to be released, contains some 500 persons. For each person, four samples per channel of data are available. The first sample of the first session is used to build a biometric template. The second sample of the first session is used as a query to generate a genuine user match score of the first session whereas the two samples of the second session are used in a similar way to generate two genuine user match scores. A *template* is the data sample used to represent the claimed identity whereas a *query* is the sample with which the template is compared. The impostor scores are produced by comparing all four samples originating from another population of persons excluding the reference users.

It is important to distinguish two data sets, i.e., the *development* and the *evaluation* sets. The development set is used for algorithm development, e.g., finding the optimal parameters of an algorithm, including setting the global or client specific decision threshold. An important distinction between the two sets is that the population of users in these data sets are *disjoint*. This ensures that the performance assessment is unbiased. There are 51 genuine users in the development set and 156 in the evaluation set. These two sets of users constitute the 207 users available in the database. The remaining 126 subjects are considered as an external population of users who serve as zero-effort impostors.

The *development impostor score set* consists of two score sets of equal size, each having $103 \times 4$ samples, i.e., 103 persons and each contributes 4 samples. These two score sets were used in conjunction with the first- and second-session genuine match scores, respectively. This design ensures that the impostors used in Sessions 1 and 2 are not the same. For instance, when training a fusion algorithm, Session 1 data can be used to learn the parameters of the fusion classifier and Session 2 data can be used as a validation data set. Such a characteristic is important for algorithm development.

The *evaluation impostor score set* also further consists of two data sets, each having 51 and 126 subjects set apart as zero-effort impostors, used for Session 1 and Session 2 data, respectively. Note that the 51 "impostor subjects" of the Session 1 evaluation set are actually genuine users in the development data set. This does not contribute any systematic bias when measuring the performance because the genuine users are disjoint in both the development and evaluation data set. The motivation for using two different impostor populations in the evaluation data sets for Sessions 1 and 2 again is to avoid systematic and optimistic bias in assessing *client-specific* fusion classifiers trained on the Session 1 data. Unlike the common fusion classifiers reported in the literature [1] (and references therein), client-specific fusion classifiers adapt themselves to each genuine user or claimed identity. Under our experimental design, these classifiers can further use the Session 1 evaluation data for training but must use the Session 2 evaluation data for assessing the performance. Having two different sets of impostors will thus avoid

TABLE IV
EXPERIMENTAL PROTOCOL FOLLOWED IN ORDER TO GENERATE MATCH SCORES FROM THE BIOSECURE DS2 DATABASE. $S1/S2$ = Sessions 1 AND 2

| Data sets | | No. of match scores per person | |
|---|---|---|---|
| | | dev. set (51 persons) | eva. set (156 persons) |
| S1 | Gen | 1 | 1 |
| | Imp | $103 \times 4$ | $51 \times 4$ |
| S2 | Gen | 2 | 2 |
| | Imp | $103 \times 4$ | $126 \times 4$ |

$\cdot \times \cdot$ denotes persons $\times$ samples. This number should be multiplied by the number of persons in the above set (e.g., 51 for development set) in order to obtain the total number of accesses for the genuine or the impostor classes.

the situation where a fusion classifier is tested on impostor data on which it has already been trained to distinguish the impostors.

Table IV shows the partitioning of the genuine user and impostor score sets of the development and evaluation data. The exact number of accesses differs from that listed in this table because of missing observations caused by the failure of the segmentation process or other stages of the biometric authentication system. The experimental protocol involves minimal manual intervention. In the event of *any* failure, a default score of "−999" is outputted. Similarly, a failure to extract quality measures will result in a vector containing a series of "−999."

Although the desktop scenario involves supervised data acquisition, the level of supervision differs from one collection site to another. As a result, there exists a site-dependent bias in terms of performance and this bias is readily observable from the captured images for face and fingerprint biometrics.

In Sections IV-A–C, we shall explain the two evaluation schemes.

### A. Cost-Sensitive Evaluation

The cost-sensitive evaluation was designed with two goals:
1) to assess the robustness of a fusion algorithm when some match scores and/or quality measures are not present; this is typically due to failure-to-acquire and/or failure-to-match;
2) to test how well a fusion algorithm can perform with minimal computation and hardware cost.

Note that a "cost" can also be associated with the time to acquire/process a biometric sample. Hence, longer time implies higher cost, and vice versa.

Assigning a cost to a channel of data is a very subjective issue. In this study, we adopt the following rules of thumb:
1) If a device is used at least once, a fusion algorithm will be charged a unit cost, although we are aware that in reality, different devices may have different cost. This choice is clearly device and task dependent.
2) The subsequent use of the same device will be charged 0.3 of a unit in view of the fact that the same hardware is being reused.
3) A device is considered used if a fusion algorithm acquires a sample for subsequent processing, i.e., to extract quality measures and/or to obtain a match score. This is regardless of whether the resulting match score will actually contribute to the final combined score.

Through the cost-sensitive evaluation, the design of a fusion algorithm becomes more challenging because the task now is to maximize the recognition performance *while* minimizing the cost associated to the device usage. In this respect, there exists two strategies to solve this problem, which can be termed as a *fixed parallel* and a *sequential* approach. A fixed parallel solution preselects a set of channels and use them for all access requests. A sequential solution, on the other hand, may use different channels for different access requests. The sequence of systems used is determined dynamically.

### B. Cross-Device Quality-Dependent Evaluation

The goal of this evaluation experiment is to assess the ability of a fusion algorithm to select the more reliable channels of data, given quality measures derived from biometric data. The task is made more challenging with cross-device matching, i.e., a matching can occur between a biometric template acquired using one device and a query biometric data acquired using another device. In our case, the template data is always acquired using a high quality device (giving better verification performance) and the query data may be acquired using a high or a low-quality device. Note that cross device matching occurs only in the latter case. The channels of data considered are face and the three right fingerprints, denoted as $fnf$, $fo1$, $fo2$, and $fo3$. In case of cross device matching, these channels are denoted as $xfa$, $xft1$, $xft2$, and $xft3$. The development set consisting of scores and quality measures corresponding to all eight channels were distributed to the participants. The (sequestered) evaluation set, on the other hand, contains only four channels of data as a result of mixing $fnf/xfa$ (face taken with a digital camera/webcam) and $fo\{n\}/xft\{n\}$ for all $n \in \{1, 2, 3\}$ (optical/thermal fingerprint sensor for three fingers; see description in Table II). These four channels of data can be any of the following combinations:

1) $[fnf, fo1, fo2, fo3]$—no device mismatch;
2) $[fnf, xft1, xft2, xft3]$—device mismatch for the fingerprint sensor;
3) $[xfa, fo1, fo2, fo3]$—device mismatch for the face sensor;
4) $[xfa, xft1, xft2, xft3]$—device mismatch for both the face and fingerprint sensors.

A fusion algorithm does not know from which device a biometric sample was acquired since the identity of the device is unknown. This is a realistic scenario because as a biometric technology is deployed, it may be replaced by a newer device. Furthermore, its configuration may change, resulting in its acquired query biometric data being significantly different from the previously stored template data. This fusion problem is challenging because each of the four combinations require a different fusion strategy in order to achieve the optimal result.

### C. Simulation of Failure-to-Acquire and Failure-to-Match Scenarios

For each of the above-mentioned two evaluation schemes, we also introduce a variation of the problem in order to simulate failure-to-acquire and failure-to-match scenarios. The motivation is to evaluate the robustness of a multimodal biometric system with respect to both types of failures. In principal, a multimodal system contains redundant subsystems, each of which produces a hypothesis regarding the authenticity of an identity claim. However, to our knowledge, such redundancy has never been formally evaluated.

In order to simulate the failures, one can assume that they are device- and subject-dependent; device- and subject-independent; device-dependent but subject-independent; and, device-independent but subject-dependent. Among these four cases, we opted for the one that are both device- and subject-independent, i.e., the failures can happen randomly and spontaneously. This is actually a more difficult scenario among the four, as the failures are completely unpredictable. If they were, one could devise the following solutions: replace a particular device that is malfunctioning in the device-dependent case, or recommend a user to use a different biometric modality in the subject-dependent case. If a fusion algorithm can withstand our chosen scenario, the remaining three scenarios can, therefore, be solved easily. Based on this rationale, we shall focus on the device- and subject-independent case.

We shall introduce missing values only on the evaluation data set, and *not* the development data set. The reason is that the development data set is often better controlled. The missing values are introduced for each of the genuine or impostor match scores *separately* as follows: Let $M$ be a matrix of scores of $N$ samples by $d$ dimensions (corresponding to all the $d$ columns of match scores from $d$ devices: face, 6 fingers, and 1 iris). The total number of elements in $M$ is $d \times N$. Missing values were gradually introduced by replacing $T$ observed values with "$-999$" (the dummy value denoting missing value) in such a way that all the elements in the matrix $M$ have equal probability of being deleted. We varied $T$ such that the ratio of $T/(dN)$ was 10%, 20%, 30%, and 40% and that the subsequent subset always contained missing values of its precedent subset.

## V. EXPERIMENTAL RESULTS

Two types of errors can occur in biometric authentication: false acceptance and false rejection. Both errors are quantified by their respective error rates: false acceptance rate (FAR) and false rejection rate (FRR). These rates are obtained by counting the respective event (false acceptance or rejection) for a given threshold. We use two derived indicators from these two measures, namely, EER and half total error rate (HTER). EER is defined as the operating point where FAR is equal to FRR, and there is only one unique threshold (by means of interpolating the FAR and FRR curves, if necessary) satisfying this condition. This rate is calculated by counting the number of false acceptance (resp. false rejection) instances when the claimant is an impostor (resp. a genuine user/client/enrollee) normalized by the total number of impostor (resp. client) accesses. HTER, on the other hand, is the average of FAR and FRR. In all the evaluations, the particular threshold leading to the reported HTER is supplied by the participants. This threshold was determined by satisfying the EER constraint (FAR equals FRR) on the development data set. Hence, although a fusion system may have a very low EER, due to badly estimated threshold, its HTER value can still be relatively very high.

Sections V-A–C present the results of the two evaluation schemes.

### A. Cost-Sensitive Evaluation

In the cost-sensitive evaluation, the submitted fusion systems can be divided into three types: fixed static fusion, exhaustive fusion, and sequential fusion. Fixed static fusion uses only a subset of fixed biometric systems identified using the development data set provided. Exhaustive fusion is a special case of fixed static fusion that uses *all* the available subsystems (eight in our case). Sequential fusion, on the other hand, uses a variable number of systems. The order in which the subsystems are evaluated as well as the upper and lower thresholds (used to assess the level of confidence of a decision) are determined on the development data set (refer to Section II-C). Fig. 1(a) shows the performance (assessed on the evaluation set) of the baseline (unimodal) systems (in blue), sequential fusion (green), fixed static Naive Bayes fusion with two sensors (red) and three sensors (magenta), and exhaustive fusion (black).

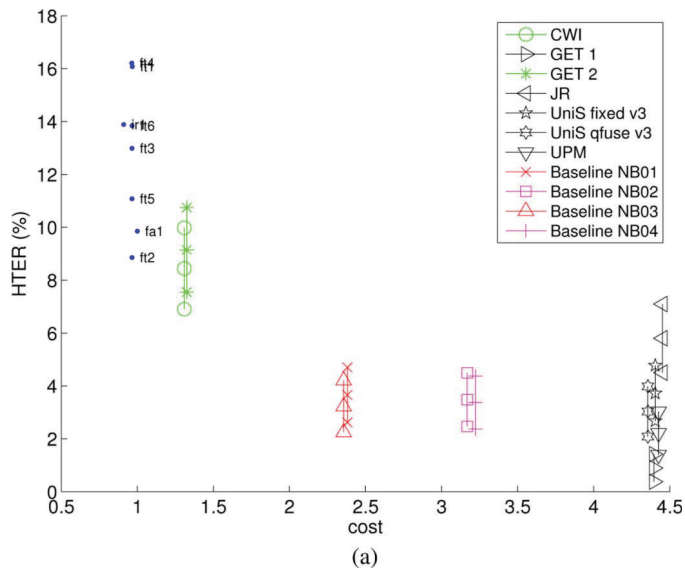

Fig. 1. Performance of all submitted fusion systems participating in the cost-sensitive evaluation (each bar showing the upper and lower confidence interval in terms of HTER, assessed on the evaluation set). The performance of each fusion system is plotted along with their confidence interval calculated according to [44].

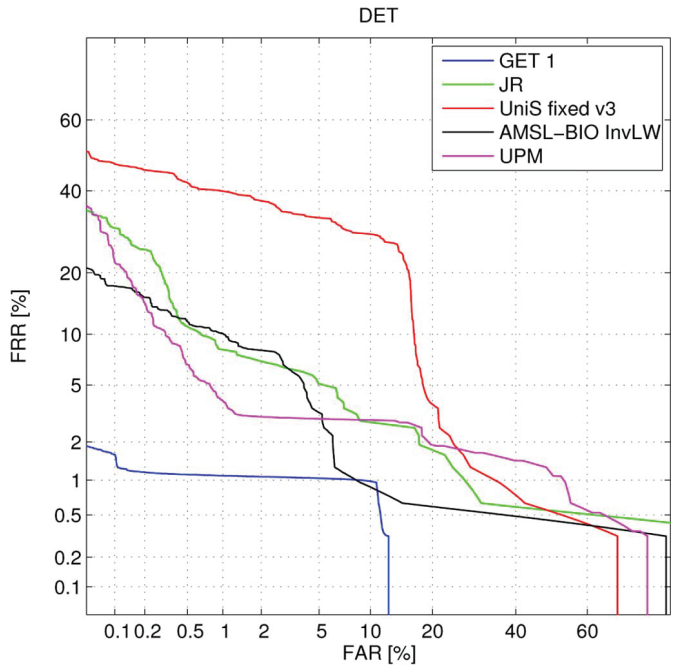

Fig. 2. Performance of exhaustive fusion systems.

*1) Baseline Systems:* The performance of the baseline (unimodal) systems exhibit somewhat higher error rates than those recently reported in the literature. This is because first of all, no effort has been made to fine tune the baseline systems. For instance, contrary to what is commonly reported in the literature, our iris baseline system has much higher error rate, due to the suboptimal iris segmentation algorithm as well as relatively uncontrolled iris acquisition procedure (e.g., blurred images were found). The second reason is that no data is discarded. As a result, failure-to-acquire and failure-to-match errors have already been considered when calculating the error. Recall that the match scores as a result of using the erroneous samples are assigned a dummy value of "999"—a value low enough to cause a rejection decision in *all* cases.

*2) Fixed Static Fusion Systems:* The performance of the fixed static fusion systems, shown in red and magenta in Fig. 1, is that of the Naive Bayes classifiers (assuming independence) using biometric subsystem outputs. The first group in red consists of face and fingerprint systems (with two fingers), resulting in a cost of $1 + 1 + 0.3 = 2.3$. The second group in magenta has a cost of 3.3, as a result of the additional iris subsystem. The generalization error measured in terms of HTER (for a fixed decision threshold) of the second group is certainly not significantly better than that of the first group.

*3) Exhaustive Fusion Systems:* On the bottom right of Fig. 1 are the exhaustive fusion systems (with the maximal cost of 4.5 unit). While the best system within this group (plotted in black color) did indeed achieve the lowest generalization error, most of the systems are only marginally better than the two fixed static fusion systems.

The task of designing a fusion classifier transforms to the one of choosing the best fusion candidate for a given cost on the development set in the hope that the chosen fusion candidate can generalize on the (sequestered) evaluation set. Since there is a mismatch between the development and evaluation set in terms of population of enrolled (genuine) users as well as impostors, and their respective sizes (hence a change in class priors), the design of a fusion classifier was fairly challenging.

In order to optimize the performance for a given cost, one typically has to consider all possible combinations of fusion candidates. Doing so requires one to perform $2^8 - 1 = 255$ combinations of fusion candidates (minus one for the empty set), and the 255 combinations can have only 18 unique costs, ranging from 1 (considering only one channel of data) to 4.5 (all eight channels). In the reported evaluation exercise, the participants had the right to submit as many fusion systems/candidates as possible. However, no explicit request was made to the participants to provide a solution for *each* of the possible costs. Doing so would end up in 18 solutions for each participating team—an impractical proposition from the organizer's view point. It turned out that each participating team in the end provided only one fusion system optimizing the most expensive cost, i.e., the exhaustive fusion system (with 4.5 unit of cost).

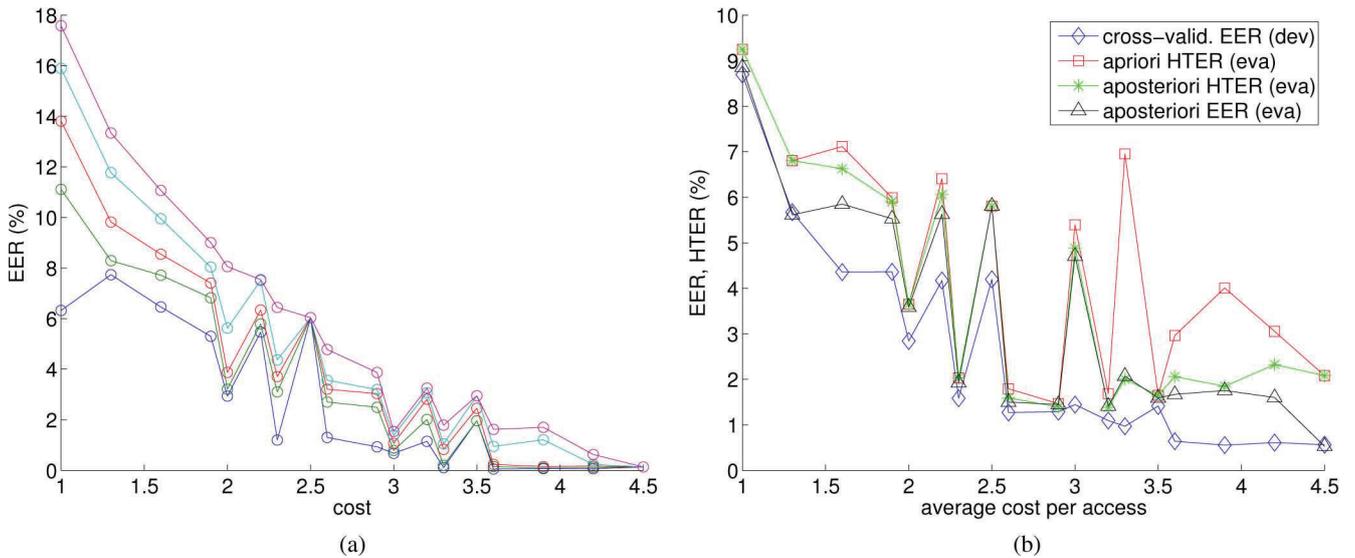

Fig. 3. (a) Optimization of UniS GMM-Bayes fusion classifier by a two-fold cross validation on the development set. (b) Rank-one performance versus average access cost. This GMM-Bayes system was provided by the organizer.

Although the UPM and JR submissions contain some channel selection, the selection process makes use of the quality information. Even if the acquired sample is not subsequently used for matching (due to its low quality), both methods are considered exhaustive fusion.

In order to obtain HTER, each participant had to provide a decision threshold. As can be observed, some of the submitted decision thresholds were not optimal. For instance, the threshold of the AMSL-BIO InvLW system was so suboptimal that it resulted in 50% HTER.

Given correctly estimated decision thresholds, the remaining exhaustive fusion algorithms can now be compared on equal grounds. The top three solutions are GET-1, UPM, UniS qfuse. These three systems share a common characteristic: they are trainable fusion classifiers. GET-1 is a GMM-Bayes classifier, hence generative in nature. UPM and UniS qfuse are discriminative classifiers, both based on their own implementation of logistic regression. In theory, both generative and discriminative approaches are equally powerful, given correctly estimated parameters. In practice, however, their performance may differ due to different implementation, resulting in different estimates of model parameters. The analysis of the DET curves of the exhaustive systems show that for this data set, GET-1 is the best fusion classifier, as shown in Fig. 2.[2]

*4) Post-Experimental Analysis:* The exhaustive fusion, in general, has the best performance (lowest HTER) among the three different types of fusion strategies. However, it also entails a higher average cost per access. Such a trend is not readily observable in Fig. 1. As the organizer (UniS) of the competition, we shall introduce a postexperimental analysis, analyzing performance spanning all possible fusion candidates.

The GMM-Bayes fusion classifier was trained on the entire score feature space (a total of eight dimensions). It was then tested on all the 255 combinations of the score feature space by means of Gaussian marginalization (as discussed in Section II-B1). Missing values were handled in the same way. For instance, if features 1, 2, and 4 are chosen, and 4 is missing, then the UniS GMM-Bayes will calculate the final combined score using only features 1 and 2.[3]

The GMM-Bayes classifier was preferred over the Naive Bayes classifier for the postexperimental analysis (as reported in Section V-A2) because there is a *strong correlation* among the genuine match scores of different fingers of the same subject (as high as 0.6), whereas the correlation among different biometric modalities (e.g., face versus iris) is close to zero. In all cases, no correlation is observed for the impostor match scores. The GMM-Bayes fusion classifier is thus better suited for these 255 fusion tasks.

In order to estimate the fusion performance using only the development set (recalling that the evaluation scores were sequestered), we employed a two-fold cross-validation. The resultant performance, measured in terms of averaged EER of the two folds, across all 255 combinations, is shown in Fig. 3(a). Plotted in this figure are the median (red), the upper, and lower quantiles (cyan and green lines respectively), and, the upper and lower range (in purple and blue lines) of performance in HTER for a given cost. Note that there is only one possible way to obtain a cost of 2.5, i.e., by combining all six fingers, hence explaining the convergence of performance to a single point. Among these curve, the lowest one (in blue) is the most important one because the goal here is find a fusion candidate that has the lowest error at a given cost.

The performance versus cost curves presented in Fig. 3(b) are called "rank-one" cost-performance curve. This means that only the performance of the best fusion candidate (using the GMM-Bayes classifier) is reported. In a rank-two curve, one would choose the minimum of the top two performing candidates to plot the curve, etc. Three of the four curves were computed on the evaluation set and only one on the development set. The latter is plotted here (in blue) in order to show the actual performance optimized on the development set via the two-fold cross validation. The reported error is the average EER of the two folds. The EER measure (rather than HTER) is more suitable in this context so that the performance is independent of the choice of the decision threshold. The remaining three curves are explained below.

[2]The DET curve of AMSL-BIO InvLW is included for comparison. Although its HTER was evaluated to be 50% due to bad threshold setting, the actual EER would be around 4.5%. Also the DET curve of UniS-qfuse could not be plotted because its genuine match scores contain only three values less than 0.0002, nine values more than 0.9101, and in between these two values found $\{0.0937, 0.3637\}$. Such discontinuity makes it virtually impossible to visualize the DET curve.

[3]Due to the way the missing values are handled, this classifier is different from the GET system.

1) *a priori* **HTER**: This rank-one curve (plotted in red) shows the *achievable* generalization performance if one were to use the fusion candidates minimizing a given cost, based on the development set, via cross-validation.
2) *a posteriori* **HTER**: This rank-one curve (plotted in green) shows the actual performance in terms of HTER of the fusion candidate on the evaluation set. The assumption here is that the evaluation set is available but the optimal decision threshold is unknown.
3) *a posteriori* **EER**: Finally, this rank-one curve (plotted in black) is similar to the previous one, reporting the performance of the fusion candidate optimizing a given cost on the evaluation set, except that it also assumes that the optimal threshold is known. This curve is hence reported in EER.

When optimizing a fusion classifier without any knowledge of the evaluation set (using the sequestered (held out) data set scenario), the best performance one can obtain is the first (*a priori*) curve. This curve is directly comparable with the performance submitted by the participants (shown in Fig. 1).

The second and third (rank-one) curves are not achievable; they are shown here in order to show the oracle cases, where the evaluation set is available for the second curve; and on top of that, the optimal decision threshold is known for the third curve. As can be observed, by injecting more information, the error actually decreases from the first to the second curve; and, from the second to the third curve.

These curves show that the actual achievable fusion performance is dependent on two factors: the fusion candidate and the (correct) decision threshold. Choosing the correct candidate given only the development set requires a criterion yielding a solution that can generalize well across populations. In [45], the authors demonstrated that such a criterion can be effectively realized using parametric error bounds such as the Chernoff and Bhattacharya bounds [28], rather than computing the EER of the fusion performance empirically, as commonly practised. Error bounds, however, do assume that the underlying scores are normally distributed and therefore, preprocessing is recommended to ensure the conformity of the data to this assumption. In practice, it was observed in [45] that even if the underlying multivariate distribution is not strictly Gaussian (as measured by the standard Gaussianity tests), the estimated bound is still better (in terms of rank-one performance-cost curve) than the empirical estimates of error (via cross-validation on the development set) for fusion candidate selection.

*5) Further Analysis of Dynamic Fusion:* This section analyses the robustness of the dynamic fusion systems. According to Fig. 1, the two sequential fusion classifiers (in green), i.e., CWI and GET-2, are only slightly better than the unimodal subsystems. However, taking the cost into consideration, as these two systems rely, most of the time, on a single modality (hence giving a cost of slightly more than one unit), the solution they offer is preferable.

[4]The data set with "0%" missing value is the original data set, which itself actually contains some missing values.

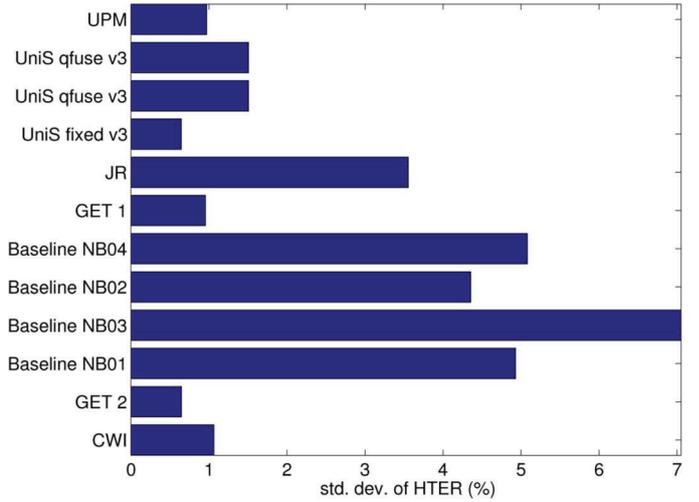

Fig. 4. Performance variation, in terms of standard deviation of HTER, for 0% to 40% missing observations, simulating failure-to-acquire and failure-to-match scenarios.

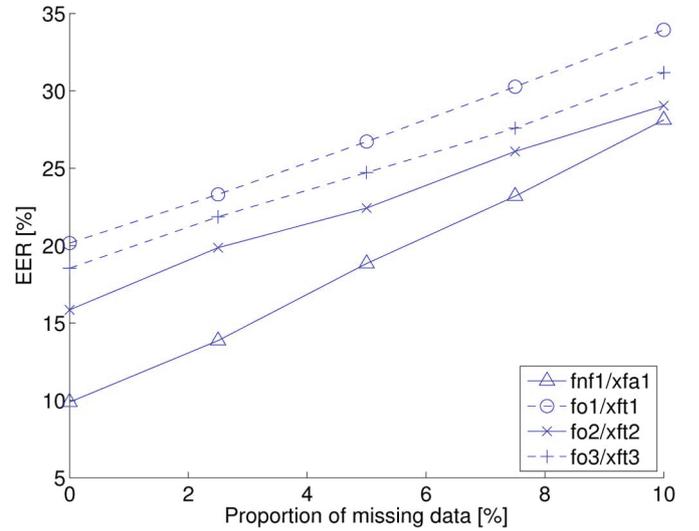

Fig. 5. Performance of biometric subsystems.

The real advantage of the two sequential fusion classifiers can be more readily appreciated under failure-to-acquire and failure-to-match, which are simulated by deleting entries, as described in Section IV-C. Fig. 4 shows the performance of different fusion systems on data sets with 0%–40% missing values given in terms of standard deviation of HTER.[4] In order to measure the variation in performance, we also measured the variance of performance of each system on the five data sets (with missing values). The result is shown in Fig. 4(b). As can be observed, despite the missing information, the performance of CWI and GET-2 remains very stable, with only ±1% of HTER. Comparatively, the unimodal systems vary by ±6% of HTER under the same scenario.

With missing values, systems with a low number of subsystems, such as the fixed static systems (the baseline Naive Bayes systems), fluctuate in performance, i.e., with at least ±4% of HTER. In comparison, most of the exhaustive fusion algorithms have performance fluctuation of around $\pm 1.5\%$ of HTER, with the exception of the JR system, which has $\pm 3.8\%$ of HTER.

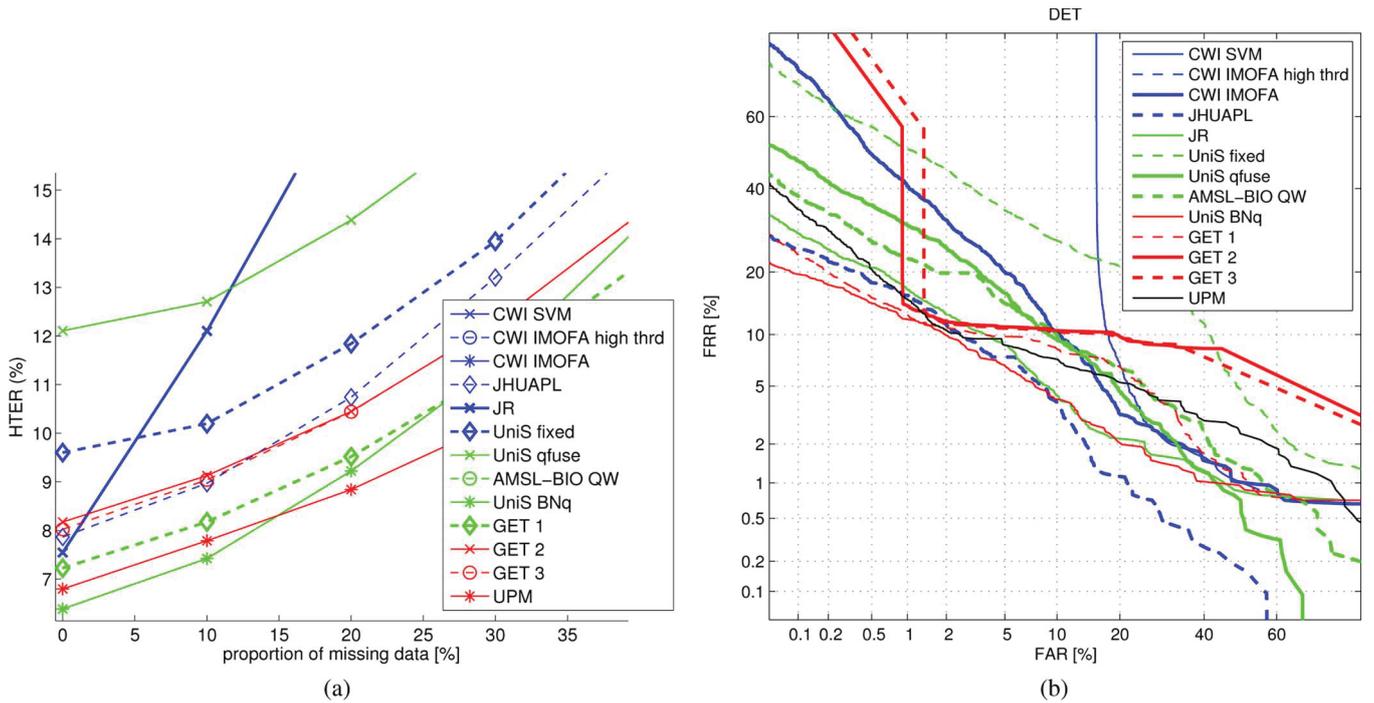

Fig. 6. While (a) shows the performance of fusion system in HTER (with *a priori* chosen threshold on the provided development set) when some data is missing, (b) shows only the DET curves when no data is deleted.

### B. Cross-Device Quality-Dependent Evaluation

For the cross-device quality dependent evaluation, we first assessed the baseline performance of the four channels of data, i.e., $fnf1/xfa1$, $fo1/xft1$, $fo2/xft2$, and $fo3/xft3$, where $fnf1/xfa1$ means that in this channel of data, the query face images are captured by the digital camera ($fnf1$) or a web-cam ($xfa1$). The $fnf1$ channel has a higher image quality than the $xfa1$ channel. The template images were captured using the digital camera. Matching the template with the $xfa\{n\}$ is considered a cross-device matching. Recall that $fo\{n\}$ means fingerprint images captured using an optical sensor whereas $xft\{n\}$ means the same subjects, but the data acquired using a thermal sensor by sliding a finger over the sensor. An ideal fusion system should consider from which device the query images are captured and use the right fusion strategy given only scores and quality measures.

Fig. 5 shows the baseline performance of the biometric subsystems (prior to fusion). Similar to the cost-sensitive evaluation, here, an increasing number of entries of the data are deleted, resulting in the proportion of missing data between 0% to up to 40%. The access requests with missing observations of scores/quality measures are automatically rejected by the system. As a result, as the proportion of missing data is increased, more and more false rejections of genuine accesses occur, resulting in increased EER.

The fusion performance of the submitted systems is shown in Fig. 6(a). In this assessment, all the observable channels of data are used, contrary to the cost-sensitive evaluation. Our focus here is to assess how well a fusion system can perform with changing image quality and in the presence of an increasing number of missing observations. As can be observed, the top two systems are UniS BNq and UPM. These systems are device-specific, meaning that they first estimate how probable the channel of data is from the observed quality measures and then use the corresponding device-dependent fusion function. The next best system is GET-1, which is a Bayesian classifier whose class-conditional densities are estimated using a mixture of Gaussian components. This system does not use the quality measures provided and so does not change its fusion strategy under cross-device matching. Some of the systems that actually use the quality measures are not among the best systems because they did not use the right threshold. Such is the case for JHUAPL. To assess the performance independently of decision threshold, we plotted the DET curves of all the systems with the original data set (prior to introducing any missing data) in Fig. 6(b). Here, the JHUAPL fusion system performs very well for low FRRs whereas UniS-BNq dominates for low FARs. The winning systems are those that exploit the quality information. This experiment shows that quality measures can be used to mitigate the effect of cross-device mismatch in the context of multimodal biometrics.

### C. Future Research Issues

The following are some important future research issues to be considered:

1) **Quality-based fusion within a single device**: The current benchmarking effort is limited to the context of varying levels of biometric quality due to cross-device matching. Other factors that can affect the quality of biometric samples are the user interaction and the acquisition environment, as exemplified by Alonso *et al.*'s study for the fingerprint modality [46]. A recent work [47] demonstrated that, indeed, exploiting the quality variation *within* a single device, as well as the variation *between* devices can improve the generalization performance under cross device matching much more significantly

2) **More quality measures**: The current data set is limited in the number of quality measures available. A direct extension is to introduce more quality measures, especially the fingerprint ones, e.g., [46]

3) **Sequential fusion**: The potential of sequential fusion has not been fully explored. The current strategy to select the next most informative fusion candidate is very heuristic. Furthermore, a principled method for the selection of the upper and lower thresholds (determined by the level of desired confidence) is also needed.

## VI. CONCLUSION

The BioSecure DS2 multimodal evaluation campaign aimed at assessing fusion algorithms under restricted computational resources, possible hardware failures, and changing image quality. For the purpose of evaluation, a data set consisting of match scores and quality measures was carefully designed in order to benchmark multimodal fusion algorithms. This data set was constructed for the application of physical access control of a medium-sized establishment (from some 300 to 500 users). This campaign gathered 22 fusion algorithms, all evaluated using the same generated data set. The evaluated fusion algorithms are very diverse, including generative classifiers, e.g., Bayesian belief network, Bayesian classifiers, and Dempster–Shafer theory of evidence; discriminative classifiers, e.g., SVMs and logistic regression; and transformation-based classifier combiners (where biometric subsystem outputs are normalized individually and then combined using fixed fusion rules such as sum or product).

Our findings suggest that, while using all the available biometric sensors can definitely increase the fusion performance, such benefit comes at the expense of acquisition time, computation time, and the cost associated with installing the physical hardware and its maintenance cost. A promising solution which does not increase this compounded cost is sequential fusion, as demonstrated in our experiments. A sequential fusion strategy only outputs the final combined match score as soon as sufficient confidence is attained, or when all the match scores available are exhausted. It saves computational cost because for most access requests, only a few biometric subsystems are needed. In the presence of changing image quality, which may be due to change of acquisition devices and/or device capturing configurations, we observe that the top performing fusion algorithms are those that exploit the automatically extracted biometric trait quality measures in order to identify the most probable biometric device from which the query biometric data was derived.

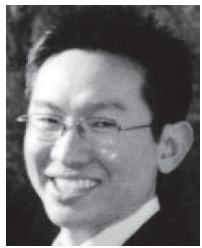

**Norman Poh** received the Ph.D. degree in computer science from the Swiss Federal Institute of Technology in Lausanne (EPFL), Switzerland, in 2006.

He benefitted from two successive personal research funding grants from the Swiss NSF (2006–2010) to investigate adaptive learning of multiple classifier systems at the Centre of Vision, Speech and Signal Processing (CVSSP), University of Surrey (UniS). He is also a work-package leader in the EU-funded Mobile Biometry (MOBIO) project at UniS.

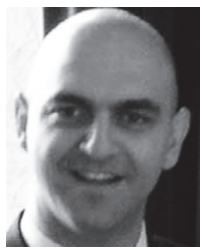

**Thirimachos Bourlai** received the Ph.D. degree in electrical and computer engineering from the University of Surrey, U.K., in 2006. There he continued working with Prof. J. Kittler as a postdoctoral researcher until August 2007.

He then worked as a Research Fellow in a joint project between the University of Houston and the Methodist Hospital, TX. Since February 2009 he is a Visiting Research Assistant Professor at the Biometrics Center, West Virginia University. His areas of expertise are image processing, pattern recognition, and biometrics.

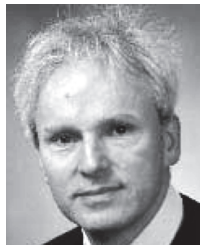

**Josef Kittler** graduated from the University of Cambridge in electrical engineering in 1971, where he also received the Ph.D. degree in pattern recognition, in 1974, and the Sc.D. degree, in 1991.

Since 1986, he is a Professor in charge of the Centre for Vision Speech and Signal Processing, University of Surrey, U.K. He published more than 200 papers, coauthored a book, and is a member of the Editorial Boards of various IEEE Transactions. His current research interests include pattern recognition, neural networks, image processing, and computer vision.

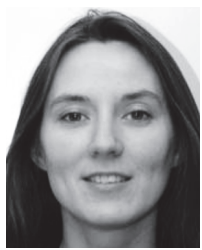

**Lorene Allano** received the Ph.D. degree in computer science from the Institut National des Télécommunications, France, in 2009, for her work on multi-biometrics.

She then moved to Commissariatà L'énergie Atomique, France (CEA) within the Laboratoire Intelligence Multi-capteurs et Apprentissage (LIMA Laboratory). Her areas of expertise are multiple classifier systems, pattern recognition, data mining, and biometrics.

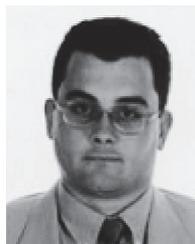

**Fernando Alonso-Fernandez** received the M.S. degree (with distinction) and the Ph.D. degree (*cum laude*), both in electrical engineering, from the Universidad Politecnica de Madrid (UPM), Spain, in 2003 and 2008, respectively.

He is currently the recipient of a Juan de la Cierva postdoctoral fellowship of the Spanish Ministry of Innovation and Science. He has published several journal and conference papers and he is actively involved in European projects focused on biometrics. He has participated in the development of several systems for a number of biometric evaluations and he has been an invited researcher in several research laboratories across Europe.

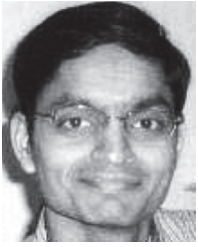

**Onkar Ambekar** received the B.E. degree in instrumentation from B.A.M. University, India, in 2000, and the M.S. degree in sensor system technology from the University of Applied Sciences in Karlsruhe, Germany, in 2004. He is currently working toward the Ph.D. degree at the CWI Institute in Amsterdam.

He has worked in Germany and Austria on object detection and tracking for vehicle guidance, and holds a patent in this area.

**John Baker** received the M.S.E.E. degree from Johns Hopkins University in 1994.

He is a member of the Johns Hopkins University Applied Physics Laboratory, where he is engaged in research and development of biometric sensing and fusion methods.

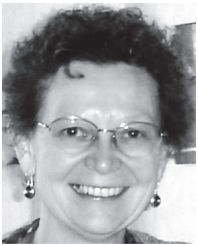

**Bernadette Dorizzi** received the Ph.D. degree (Thèse d'état) in theoretical physics from the University of Orsay (Paris XI-France) in 1983.

She is Professor at TELECOM & Management SudParis (Institut TELECOM) since September 1989, and head of the Electronics and Physics Department since 1995, where she is in charge of the Intermedia (Interaction for Multimedia) research team. She is coordinator of the BioSecure Association.

**Omolara Fatukasi** received the Ph.D. degree in electrical and computer engineering from the University of Surrey, U.K.

Her areas of expertise are in multimodal biometric fusion and pattern recognition.

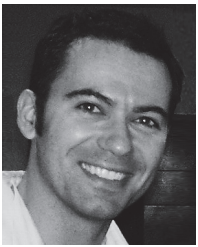

**Julian Fierrez** received the Ph.D. degree in telecommunications engineering from the Universidad Politecnica de Madrid, Spain, in 2006.

Since 2002, he is affiliated with Universidad Autonoma de Madrid in Spain, where he currently holds a Marie Curie Postdoctoral Fellowship, part of which has been spent as a visiting researcher at Michigan State University in the U.S.

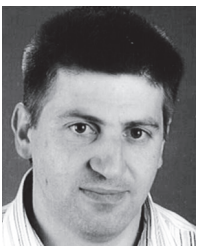

**Harald Ganster** received the Ph.D. degree with distinction in technical mathematics from the University of Technology Graz, Austria, in 1999.

He is currently with the Institute of Digital Image Processing at Joanneum Research in Graz and is responsible for the field of biometrics.

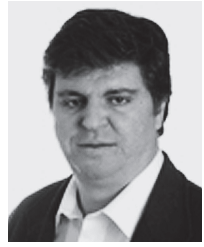

**Javier Ortega-Garcia** received the Ph.D. degree (*cum laude*) in electrical engineering from Universidad Politecnica de Madrid, Spain, in 1996.

He is founder and codirector of the Biometric Recognition Group—ATVS. He is currently a Full Professor at the Escuela Politecnica Superior, Universidad Autonoma de Madrid. His research interests are focused on biometrics signal processing. He has published over 150 international contributions, including book chapters, refereed journal and conference papers. He has chaired "Odyssey-04, The Speaker Recognition Workshop," cosponsored by ISCA and IEEE, and cochaired "ICB-09, the 3rd IAPR International Conference on Biometrics.

**Donald Maurer** received the Ph.D. degree in mathematics from California Institute of Technology in 1969.

He is a member of the Johns Hopkins University Applied Physics Laboratory, where he develops algorithms for sensor data fusion and target discrimination.

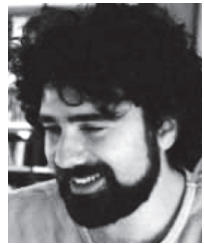

**Albert Ali Salah** received the Ph.D. degree at the Perceptual Intelligence Laboratory of Boğaziçi University.

He is currently with Intelligent Systems Laboratory Amsterdam (ISLA), at the University of Amsterdam. His research interests are biologically inspired models of learning and vision, with applications to pattern recognition, biometrics, and ambient intelligence. With his work on facial feature localization, he received the inaugural EBF European Biometrics Research Award in 2006.

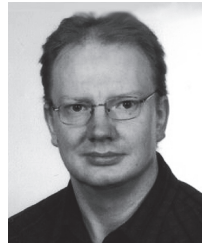

**Tobias Scheidat** received the M.Sc. degree in Computer Science in 2005 from the Otto-von-Guericke-University Magdeburg, Germany.

Since 2003, he works in the Biometric Research Group of the Advanced Multimedia and Security Laboratory (AMSL) at Otto-von-Guericke University Magdeburg. In addition to online handwriting, his research interests are focused on multibiometric fusion, biometric hashing, the evaluation of biometric systems, and meta data in context of biometrics.

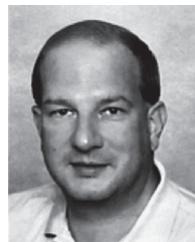

**Claus Vielhauer** received both the M.Sc. and Ph.D. degrees from Technical University Darmstadt.

is a full professor (2007) for IT Security at Brandenburg University of Applied Sciences and the leader of the biometric research group at the Advanced Multimedia and Security Laboratory at Otto-von-Guericke University Magdeburg (2003). Formerly, has been a member of the Multimedia Communications Laboratory Research Group of Prof. R. Steinmetz, Technical University Darmstadt. His main research interests cover the areas of IT and media security and forensics, biometrics and human–computer interaction and he has been performing research and management in numerous EU- and non-EU-funded research projects such as BioSecure.